# Improvement of Performance in Freezing of Gait detection in Parkinson's Disease using Transformer networks and a single waist-worn triaxial accelerometer

Luis Sigcha[a,b], Luigi Borzì[c], Ignacio Pavón[a,∗], Nélson Costa[b], Susana Costa[b], Pedro Arezes[b], Juan Manuel López[a], Guillermo De Arcas[a]

[a]Instrumentation and Applied Acoustics Research Group (I2A2), ETSI Industriales, Universidad Politécnica de Madrid, Campus Sur UPM, Ctra. Valencia, Km 7, 28031 Madrid, Spain

[b]ALGORITMI Research Center, School of Engineering, University of Minho, 4800-058 Guimar˜aes, Portugal

[c]Department of Control and Computer Engineering, Politecnico di Torino, 10129 Turin, Italy

**Abstract**

Freezing of gait (FOG) is one of the most incapacitating symptoms in Parkinson's disease, affecting more than 50% of patients in advanced stages of the disease. The presence of FOG may lead to falls and a loss of independence with a consequent reduction in the quality of life. Wearable technology and artificial intelligence have been used for automatic FOG detection to optimize monitoring. However, differences between laboratory and daily-life conditions present challenges for the implementation of reliable detection systems. Consequently, improvement of FOG detection methods remains important to provide accurate monitoring mechanisms intended for free-living and real-time use. This paper presents advances in automatic FOG detection using a single body-worn triaxial accelerometer and a novel classification algorithm based on Transformers and convolutional networks. This study was performed with data from 21 patients who manifested FOG episodes while performing activities of daily living in a home setting. Results indicate that the proposed FOG-Transformer can bring a significant improvement in FOG detection using leave-one-subject-out cross-validation (LOSO CV). These results bring opportunities for the implementation of accurate monitoring systems for use in ambulatory or home settings.

∗Corresponding author: ETSI Industriales, Universidad Politécnica de Madrid, Campus Sur UPM, Ctra. Valencia, Km 7, 28031 Madrid, Spain. Tel.: +34-91-067-7222

Email addresses: luisfrancisco.sigcha@upm.es (Luis Sigcha), luigi.borzi@polito.it (Luigi Borzì), ignacio.pavon@upm.es (Ignacio Pavón), ncosta@dps.uminho.pt (Nélson Costa), susana.costa@dps.uminho.pt (Susana Costa), parezes@dps.uminho.pt (Pedro Arezes), juanmanuel.lopez@upm.es (Juan Manuel López), g.dearcas@upm.es (Guillermo De Arcas)

1. Introduction

Freezing of gait (FOG) is one of the most disabling symptoms in Parkinson's disease (PD), characterized by brief episodes of inability to step or the presence of short steps when initiating gait, on turning while walking (Nutt et al., 2011) or when experiencing stressful situations (Nonnekes et al., 2015). FOG affects between 50% and 80% of people with PD (Weiss et al., 2015), and its presence is associated with an increased risk of falls, affecting the quality of life (Moore et al., 2007).

When a FOG episode appears, PD patients can present variability in the gait pattern, with a reduction in step length, shuffling steps, trembling of the legs, and total akinesia with a loss of movement of the limbs or trunk (Okuma, 2014). FOG episodes can have a duration of a few seconds (1 second or less for very short episodes and more than 5 seconds for long episodes) and appear more frequently during typical daily-life conditions than during straight walking assessments in clinical and laboratory settings (Okuma, 2014; Nonnekes et al., 2015).

FOG assessment involves the identification of the presence (or absence) of FOG episodes and also aims to identify their severity (Mancini et al., 2019). Assessing FOG in the clinical practice is difficult because of the lack of an optimal freezing score, and difficulties related to the clinical assessment often performed on conditions that hinder the appearance of FOG events during evaluation; for example, the assessment is usually made in the ON state, while FOG occurs more often in OFF state (Schaafsma et al., 2003; Mancini et al., 2021).

Although the clinical assessment provides relevant indicators for the characterization of FOG, the conditions whereby these are performed do not accurately represent the severity of FOG in daily life (Rahman et al., 2008; Snijders et al., 2008), such as the patients' homes, where FOG events tend to occur more frequently (Nieuwboer et al., 1998).

Therefore, the development of objective and reliable FOG monitoring mechanisms is necessary for the proper implementation of pharmacological treatment (i.e., levodopa treatment), as well as for the development of novel therapies to treat the FOG-associated symptomatology (Mancini et al., 2019; Nonnekes et al., 2015). Consequently, an accurate and objective assessment of FOG should be performed ideally during daily-life activities in home settings with the aim of assessing the full spectrum of the symptom, including the severity of FOG events and their fluctuations over the course of the day (Mancini et al., 2021).

Wearable devices have shown high potential for the development of systems for FOG detection in both laboratory and home settings (Silva de Lima et al., 2017; Pardoel et al., 2019). They are equipped with sensors (i.e., accelerometers or gyroscopes) that can provide objective measures for the quantification of motor symptoms (Rovini et al., 2017) such as FOG or gait disturbances, to improve the monitoring of the disease progression or the effects of the treatments (Mancini et al., 2021). In addition, the portability and low cost of wearable devices can allow the development of affordable systems for continuous tracking in an unobtrusive fashion without increasing the burden on patients (Rovini et al., 2017; Monje et al., 2019; Del Din et al., 2021).

Machine learning (ML) techniques in conjunction with wearable sensors have led to unprecedented performance in a variety of applications related to motor symptom assessment (Borzì et al., 2020b; Heijmans et al., 2019; Borzì et al., 2020a), including FOG detection (Del Din et al., 2021; Landolfi et al., 2021).

ML algorithms (i.e. neural networks, decision trees, random forest, support vector machines, etc) have provided methods for the development of FOG detection systems capable of surpassing the performance of threshold-based methods. For this task, time and frequency-domain features have been extracted from wearable sensors located in different parts of the body and used to train models in supervised, unsupervised, and semi-supervised learning approaches (Pardoel et al., 2019).

Among the machine learning algorithms, in the last years, the use of deep learning (DL) approaches has led to establishing the state-of-the-art in many domains and applications (Alzubaidi et al., 2021), such as the automatic assessment of movement disorders or the diagnosis of Parkinsonian syndromes (Mei et al., 2021) including FOG detection by using mainly data collected with inertial sensors (San-Segundo et al., 2019; Li et al., 2020; Noor et al., 2021; Naghavi and Wade, 2021; Bikias et al., 2021) and wearable devices (Camps et al., 2018; Sigcha et al., 2020).

Although the automatic detection and prediction of FOG episodes using wearable sensors have been a wide research area in the last years (Irrera et al., 2018), the development of improved FOG detection methods remains important to provide accurate long-term FOG monitoring mechanisms (Borzì et al., 2021), and for the implementation of real-time cueing systems that can help to reduce the occurrence of FOG episodes (Sweeney et al., 2019).

Challenges like unpredictability, differences in movement patterns during the occurrence of freezing episodes between OFF and ON states, and temporal resolution to distinguishing short freezing episodes, make it difficult to implement reliable automatic FOG detection systems (Pardoel et al., 2019).

Novel approaches, based on DL techniques, are promising mechanisms for the development of improved systems for FOG detection (Pardoel et al., 2019; Sigcha et al., 2020). The potential of these techniques lies in their ability to process sequential data and inertial signals collected from different sensors with a minimum pre-processing and provide similar or superior performance to classic ML algorithms (Sigcha et al., 2020). Despite the advantages provided by this technology, the use of DL approaches requires a great amount of data to obtain outperforming results (Alzubaidi et al., 2021), while the implementation of certain algorithms like recurrent neural networks (RNN) presents challenges related to the length of the sequences and expensive computational requirements (Raza et al., 2021).

Transformer networks (Vaswani et al., 2017) were introduced for sequence to-sequence learning and offer improved performance at long-range sequential modelling over RNNs. These architectures were proposed to model dependencies over the whole range of sequential data without using recurrent connections to improve efficiency (Lin et al., 2021). Transformer networks are based on attention mechanisms to analyze an input sequence and decide which parts of the sequence are important for a specific task (Shavit and Klein, 2021).

In the last years, transformers have outperformed the RNNs in sequence- based tasks such as natural language processing, computer vision, audio pro- cessing, etc. (Lin et al., 2021), and recently have been applied in human activity recognition based on inertial sensors (Shavit and Klein, 2021; Raza et al., 2021). Despite the potential advantages that this technology can provide in the analysis of sequential data, the use of transformer-based neural networks has not been evaluated for FOG detection.

## 2. Related Work

A wide spectrum of FOG detection algorithms using wearable sensors data can be found in the literature. Initially, basic threshold methods based on spectral analysis were proposed (Moore et al., 2008; Bachlin et al., 2009). To increase the performance of the detection models, ML classification algorithms were exploited in several studies (Rodríguez-Martín et al., 2017; Naghavi et al., 2019; Borzì et al., 2020; Demrozi et al., 2020).

Recently, DL approaches were proposed to detect FOG, outperforming classical ML models. As far as the wearable motion sensors concern, 3-axial accelerometers have been used either alone (San-Segundo et al., 2019; Sigcha et al., 2020; Li et al., 2020; Noor et al., 2021), in combination with 3-axial gyroscopes, (Camps et al., 2018; Bikias et al., 2021; Naghavi and Wade, 2021) or as part of multimodal systems (Shalin et al., 2021).

Different sensors' configurations were used in different studies. The number of sensors ranges from 1 (Camps et al., 2018; Sigcha et al., 2020; Borzì et al., 2020) to 9 (Bikias et al., 2021; Mazilu et al., 2013), and different sensor locations were proposed, including wrist (Bikias et al., 2021), lower back (Borzì et al., 2020; Zhang et al., 2020), waist (Camps et al., 2018; Sigcha et al., 2020), thigh (Noor et al., 2021), shank (Naghavi et al., 2019; Naghavi and Wade, 2021), or a combination of two or more positions (Demrozi et al., 2020; Li et al., 2020). Detailed literature reviews of the different proposed FOG detection methods based on wearable devices can be found in (Silva de Lima et al., 2017; Pardoel et al., 2019; Sigcha et al., 2020).

In this section, recent studies published in the last two years focusing on FOG detection using DL approaches are described.

The Daphnet dataset (Bachlin et al., 2010) was used in Li et al. (2020) and Noor et al. (2021). The database includes data from 10 PD patients while OFF therapy, of which 8 manifested FOG during the recordings, providing a total number of 237 FOG episodes. Data from three accelerometers positioned on the left shank, left thigh, and lower back were recorded while subjects performed different walking tasks.

In Li et al. (2020), raw accelerometers' readings from Daphnet dataset were segmented into 4s-long windows. Data augmentation (arbitrary rotation) was employed to balance the dataset. The classification model consisted in a combi- nation of convolutional neural network (CNN) and an attention-enhanced long short-term memory (LSTM) block. When using data from all three sensors (i.e., shank, thigh, back), accuracy 0.919, area under the curve (AUC) of 0.945, and equal error rate (EER) of 10.6% were obtained in LOSO CV. When considering only data from the back accelerometer, sensitivity 0.829, specificity 0.908, AUC 0.932, and EER 11.8% were obtained. In Noor et al. (2021), only data from the sensor placed on the thigh were analyzed. Raw accelerometer readings were segmented using 2s-long windows, with no overlap. A CNN denoising autoencoder was used, with three convolutional layers both in the encoder and decoder part of the network. Sensitivity 0.909, specificity 0.670, and accuracy 0.789 were achieved in LOSO CV.

The Cupid dataset (Mazilu et al., 2013) was used in Bikias et al. (2021). The database includes data from 18 PD patients while ON therapy, of which 11 manifested FOG during the experiments, collecting a total of 184 FOG episodes. Data from nine inertial measurement units (IMUs) attached to wrists, thighs, ankles, feet, and lower back were recorded while patients performed different walking tasks, including walking sessions with 180° and 360° turns, on wide or narrow trails with obstacles, and walking through crowded hospital rooms. In Bikias et al. (2021), data from the accelerometer and gyroscope placed on the wrist were used. Inertial signals were segmented into 3s-long windows, with 0.25s overlap. Raw signals were normalized to the maximum value and fed to a CNN with two convolutional layers and one fully-connected layer. Sensitivity 0.830 and specificity 0.880 were obtained in LOSO CV.

In Naghavi and Wade (2021), 7 PD patients were asked to walk a narrow hallway while wearing an accelerometer and gyroscope on both ankles. A to- tal number of 154 FOG episodes were registered during the experiments. Raw inertial signals were segmented into 2s-long windows, with steps of 0.25s. A one-class classifier for anomaly detection was proposed, consisting of four convolutional layers and two fully-connected layers. Sensitivity 0.630 and specificity 0.986 were achieved in LOSO validation.

In Shalin et al. (2021), 11 PD patients while OFF therapy were asked to walk a pre-defined path, including walking in a narrow corridor and turning. A total of 362 FOG episodes were recorded during the experiments. Two plantar- pressure systems were used for data acquisition. Under-sampling of non-FOG data was performed to balance the dataset. The classification model consisted in a network with 2-layers LSTM, providing sensitivity 0.821, specificity 0.895, and precision 0.253, with 95% of the total labeled FOG episodes detected.

Major limitations of the above-mentioned studies include the use of a reduced dataset (max 11 PD patients with FOG), the limited number of FOG episodes collected (max 362 FOG episodes), the PD patients' state of therapy (either ON or OFF), the laboratory setting of the experiments, and finally the use of several inertial sensors or the limited performance obtained using a single sensor.

To address some of these limitations, a system for FOG detection in home environments was proposed in Rodríguez-Martín et al. (2017). For this task, a dataset (hereinafter referred as to REMPARK-FOG dataset) was collected using a single waist-mounted triaxial accelerometer while performing activities of daily living (ADL) in their home, both ON and OFF therapy. In this study the use of time and frequency domain features were used to feed a support vector machine classifier, obtaining sensitivity 0.790 and specificity 0.747 in LOSO CV. The same authors (Camps et al., 2018) improved the classification performance using a CNN consisting of four convolutional layers and two fully- connected layers, fed with two consecutive spectral representations of the signal.

Sensitivity 0.919 and specificity 0.895 were obtained when testing the model on a subset of four patients. In a previous work (Sigcha et al., 2020), we used the REMPARK-FOG dataset (Rodríguez-Martín et al., 2017) to evaluate a CNN-LSTM detection model, fed with spectral representations of three consecutive signal windows. The algorithm achieved a sensitivity of 0.871 and specificity of 0.871 in LOSO CV. Despite the results, the margin for improvement was identified in the performance and applicability of the system intended for FOG detection in home settings.

For these reasons, this paper describes some advances in automatic FOG detection using a single waist-mounted triaxial accelerometer and a novel DL classification architecture based on the combination of convolutional and Transformers neural networks. This study was performed using data of 21 patients who manifested FOG episodes while performing ADL in home settings (Rodríguez-Martín et al., 2017).

The performance of the detection methods was evaluated using a LOSO CV, to propose a generic model which can be used with data of users that were not included during training. The results show an enhancement in FOG detection over methods proposed in the related literature (i.e., convolutional or sequential models) while maintaining a reduced temporal window.

In addition, a post-processing methodology of the predictions obtained from the automatic FOG detection system is proposed. The post-processing method- ology was developed with the aim of providing a mechanism to improve the outcomes of systems intended for long-term FOG analysis in ambulatory and free-living settings.

The remainder of this paper is organized as follows: Section 3 presents the materials and methods used in this work, including the data collection, algorithmic approaches, validation metrics and methodologies, and the post-processing methodology. Section 4 presents the results and the discussion of the proposed methodology. Finally, Section 5 presents the conclusions obtained in this study.

## 3. Materials and Methods

This section presents the materials and methods used for FOG detection. In this section is reported a description of the dataset, the signal pre-processing techniques, the ML and DL approaches used for FOG detection at the window level, and the proposed FOG detection post-processing methodology intended to analyze FOG episodes and clusters of FOG episodes.

### 3.1. FOG Dataset

The REMPARK-FOG dataset collected in Rodríguez-Martín et al. (2017) was used for evaluating the proposed methods. This dataset contains recordings from 21 patients (3 females and 18 males, 69 ±9.7 years) who presented FOG symptoms during the recording sessions, in both ON and OFF states. The participants recruited presented a mean Hoehn and Yahr (Hoehn and Yahr, 1967) scoring of 3.07 (standard deviation: 0.426) in the OFF state, while the mean FOG questionnaire (Giladi et al., 2009) was an index of 15.2 (standard deviation: 4.47).

The protocol for data collection was carried out at the patients' homes. With this free-living approach, is expected an increase in the time and frequency of the FOG episodes (Nieuwboer et al., 1998). In this protocol, patients were asked to carry out a set of scripted ADL with an approximate duration of 20 minutes. Due to the medication effect, the tests were performed in both ON and OFF states. The first test was made early in the morning without the effects of the medication (OFF state), while the second test was performed at least one hour after the patient took their usual medication (ON state). Although the protocol test was based on scripted activities, the patients performed several activities including their normal ADL (i.e., going outdoors, taking a short walk, Stand Up and Go test, brushing their teeth, etc.) in a familiar environment.

The dataset was collected through a single triaxial accelerometer from a wearable IMU located at the left side of the waist (Rodríguez-Martín et al., 2013), the accelerometer's sampling rate was set to 200Hz, and the amplitude to ±6g. The occurrence of FOG episodes in the data was labeled offline by an experienced clinician using video recordings performed during the data col- lection experiments. Over a thousand FOG episodes were collected during the recordings.

The total signal duration corresponds to 18 hours (OFF state: 9.1 hours; ON State: 8.9 hours). All patients (21) manifested FOG during OFF state, while 14 patients reported FOG during ON state. The total FOG duration is 1.89 hours (OFF state: 85.1 min; ON state: 28 min). In this dataset, 785 FOG episodes were recorded in OFF state and 273 episodes in ON state. Fig. 1 shows a histogram with the duration of FOG episodes in ON and OFF states. As shown in Fig. 1, most of the FOG events have a duration of less than 10 seconds for both motor states (ON/OFF).

In the dataset, 10.5% of the accelerometer triaxial signals correspond to FOG episodes, while 89.5% correspond to the absence of FOG episodes including normal (ADL) movements. According to this distribution, this dataset can be considered as a binary unbalanced classification problem.

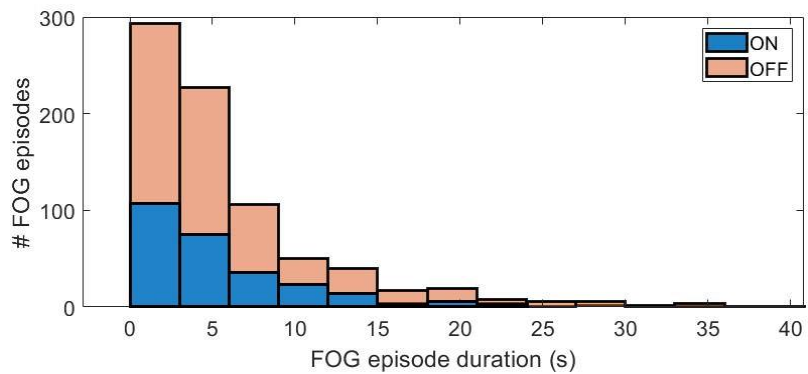

Figure 1: Histogram of FOG episodes duration in ON and OFF.

### 3.2. FOG Detection Approach

An algorithmic approach was proposed for the detection (window-based detection) and evaluation of FOG Episodes and clusters of FOG episodes. In the proposed approach, the tri-axial accelerometer signals were pre-processed (filtering and windowing) before the feature extraction. After the signal pre- processing, several feature extraction methodologies were implemented to evaluate classifiers based on ML and DL. The predictions of the best approach for FOG detection (at window level) were used to perform the post-processing analysis. In this analysis, the information of several windows was used to detect the presence of FOG episodes and clusters of FOG episodes.

The proposed approach was developed according to the scheme shown in Fig. 2. Detailed descriptions of the steps employed in the FOG detection approach are discussed in the subsequent subsections.

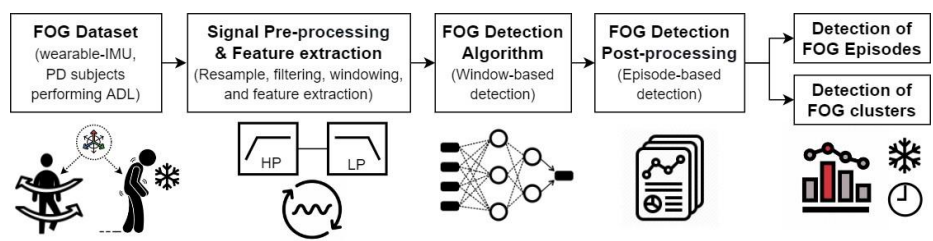

Figure 2: Algorithmic approach for FOG detection and episode detection.

### 3.3. Signal Pre-processing

The dataset comprises triaxial accelerometer signals with a sampling rate of 200 Hz. The data were resampled at 40 Hz since this frequency can be a good trade for human activity recognition using accelerometers (Khan et al., 2016). Furthermore, this sampling rate allows the analysis of typical frequencies that appear during FOG episodes (Moore et al., 2008) (i.e., freeze band from 3 to 8 Hz).

Fig. 3, shows the spectral representation of FOG episodes in both OFF and ON states. As shown in Fig. 3, most of the energy is concentrated on the freeze band during FOG events. Also, a reduction of the energy in frequencies below 3 Hz is observed.

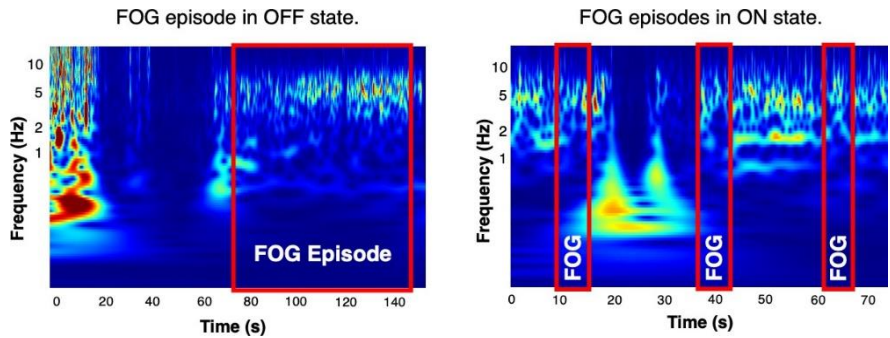

Figure 3: Spectral representation of FOG episodes in both OFF and ON states.

The resampled signals were filtered to remove the noise at high frequencies with a second-order Butterworth low-pass filter with a cut-off frequency of 15 Hz. Also, to remove the influence of gravity, a third-order Butterworth high-pass filter with a cut-off frequency of 0.2 Hz was used.

According to the related literature, the best performances in FOG detection have been obtained with the use of longer temporal windows (Moore et al., 2013; Bachlin et al., 2010), conversely, the typical FOG episodes are usually shorter than 10 seconds (Mancini et al., 2021), a situation that presents a trade-off between latency and detection performance for real-time applications.

In other related studies, the use of windows between 2 and 4 seconds have reported similar performances to those of longer windows (Mazilu et al., 2012; Zach et al., 2015). Hence, in this study, the accelerometer signals were divided into windows of 128 samples (3.2 seconds) with different overlap settings. This allows the evaluation of the algorithm's capability of being part of a cueing system intended to help the patient during FOG episodes to maintain a certain speed by using auditory, visual, or somatosensory feedback (Sweeney et al., 2019).

To assign a label to these windows, the following strategy was used: a window was labeled as FOG if more than 50% of its samples were labeled as FOG; a non- FOG window was considered only if all its samples were labeled as non-FOG. Windows containing FOG samples with less than 50% were discarded.

### 3.4. Feature Extraction Methodologies

Diverse feature extraction methodologies were reproduced to evaluate the classification algorithms for FOG detection. The sets of features include Mazilu features (Mazilu et al., 2012) (used as a baseline), the raw signals (Bikias et al., 2021), and spectral representations based on the Fast Fourier Transform (FFT) and previous windows. The spectral representations include two staked windows (Camps et al., 2018), and the use of up to 3 previous FFT windows as proposed in (Sigcha et al., 2020). The use of spectral data representations has shown high performance in FOG detection over the hand-made features using both LOSO CV and hold-out evaluations (Sigcha et al., 2020; Camps et al., 2018).

A summary of the feature extraction methodologies reproduced in this study is shown in Table 1.

Table 1: Summary of the feature extraction methodologies.

| Data representation | No. of features (input shape) | Description |
| --- | --- | --- |
| Mazilu et al. (2012) (baseline) | 21 (7*3) | Time and Frequency-domain features |
| Bikias et al. (2021) | 384 (128*3) | Raw triaxial signal |
| Camps et al. (2018) | 384 (64*6) | FFT + 1 stacked previous window |
| Sigcha et al. (2020) | 384 (2*64*3) | FFT + 1 previous window |
|  | 576 (3*64*3) | FFT + 2 previous windows |
|  | 768 (4*64*3) | FFT + 3 previous windows |

Mazilu et al. (2012) features consist of a vector of 21 features extracted from time and frequency domains (7 for each accelerometer axis), including the power of specific frequency bands and the freeze index, that corresponds to the ratio between the power of the freeze-band (3–8 Hz) and the locomotion band (0.5–3 Hz). To extract this group of features the signal was split using sliding windows with a length of 128 samples with an overlap of 75%.

The second data representation corresponds to the raw triaxial signal, as proposed in Bikias et al. (2021) as a method for FOG detection with a minimum pre-processing. The 384 features correspond to the 3 accelerometer channels with a window length of 128 each. The signals were filtered (see section 3.3) and split using sliding windows with an overlap of 10%. The data of each channel were normalized from -1 to 1 to feed the corresponding classification algorithm.

The data representation proposed in Camps et al. (2018) was composed of the symmetric part (64-bin) of a 128-point Fast Fourier Transform (FFT) extracted from each accelerometer channel. For this data representation, two consecutive windows (with no overlap) were "stacked", resulting in a 2-D representation of 384 features (64 features * 6 channels).

Also, a 3-D data representation was generated considering the addition of several previous windows as proposed in Sigcha et al. (2020). The data representation was composed of the symmetric part of the FFT (64 bin) extracted for each of the three accelerometer channels. Then, the current and the previous (FFT) windows were ordered as time steps.

The use of the current FFT window and up to 3 previous windows with different overlap settings (50% and 75%) was evaluated. Considering the current FFT and one previous window, a 3-D data representation of 384 features was obtained (2 time steps * 64 features * 3 channels). Considering the current FFT and 2 previous windows, a data representation of 576 features was obtained (3 time steps * 64 features * 3 channels). Finally, 768 features (4 time steps * 64 features * 3 channels) were obtained considering the current FFT and the three previous windows. Fig. 4 shows the process to obtain the data representation based on the FFT and previous windows.

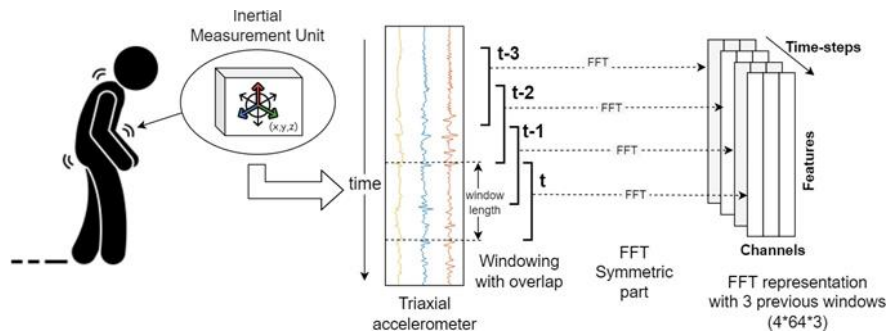

Figure 4: Fast Fourier transform (FFT) spectral representation with three previous windows (Sigcha et al., 2020).

### 3.5. Classification Algorithms

In this work, diverse classification algorithms proposed in the related literature were reproduced and evaluated. The evaluated algorithms comprise random forest (Breiman, 2001), deep neural networks (DNN) composed of convolutional layers (Bikias et al., 2021; Camps et al., 2018), and a combination of convolutional and recurrent layers (Sigcha et al., 2020). In addition, this study proposes a novel approach based on the use of Transformer blocks in combination with convolutional blocks (FOG-Transformer) as a mechanism to improve the performance in FOG detection. The architecture employed for the FOG-Transformer is shown in Section 3.5.1.

Mazilu features were evaluated with a Random Forest algorithm with 100 decision trees; this approach was selected as the baseline due to its ease of implementation and excellent results.

On the one hand, the architecture proposed in Bikias et al. (2021) that uses CNN with max-pooling and multilayer perceptron (MLP) was reproduced and evaluated with data representation based on raw signals. On the other hand, the architecture proposed in Camps et al. (2018) was evaluated with the data representation that uses FFT with one stacked previous window.

The reproduction of these two architectures (Camps et al., 2018; Bikias et al., 2021) differs from that reported in their corresponding articles in the number of sensors and signals. In both studies, the use of triaxial gyroscopes in companion with triaxial accelerometers was reported, achieving sensitivities and specificities up to 0.92 using evaluation methodologies like hold-out (Camps et al., 2018) and 10-fold cross-validation (Bikias et al., 2021). In particular, the output of the model proposed in Bikias et al. (2021) was adapted for a binary prediction (FOG or Non-FOG) instead of the reported multi-class classification (FOG, stop, walking-with-turns). Also, in both models, the loss function was changed to binary cross-entropy to make the results comparable with the other deep architectures evaluated in this study.

The recurrent architecture (CNN-LSTM) proposed in a previous work (Sigcha et al., 2020) was evaluated using the 3-D data representation that uses FFT plus 3 previous windows as shown in Fig. 4. This architecture is composed of CNN and LSTM recurrent layers, which allows analyzing sequential data from previous windows to take advantage of temporal phases that appear at the beginning of the FOG events (Cupertino et al., 2022).

The experiments to evaluate the classification algorithms were conducted on a computer with an Intel Xeon with 2.30 GHz processor, 25 GB of random- access memory (RAM), and a 12 GB NVIDIA Tesla K80 graphics accelerator. The preprocessing, and feature extraction were performed using the MATLAB software (version R2020a), while the evaluation and training of the classification models were performed in Python (version 3.6), using the libraries Keras (version 2.4), TensorFlow (version 2.3), and Scikit-learn (version 0.22).

### 3.5.1. Architecture for the Deep Neural Network with Convolutional and Trans- former Blocks (FOG-Transformer)

The FOG-Transformer model works as a classifier for FOG detection using the temporal information obtained from previous spectral windows. This architecture can work with sequential data, employing a time-distributed layer to process the time steps. The temporal behavior of the FOG episodes is analyzed using a Transformer encoder that uses attention blocks. The architecture proposed for the FOG-Transformer and its corresponding convolutional and attention blocks is shown in Fig. 5.

The FOG-Transformer is composed of three main parts, described in the following.

1. Time-distributed layer. This layer works as a wrapper for a block of 1-D convolutional layers with max-pooling and global average pooling (GAP). This block provides a mechanism to adapt the shape (3-D) of the input data to the input expected for the Transformer block. The convolutional block is used for automatic feature extraction from each single time step composed of triaxial FFT channels. In the time distributed layer, the same convolution and pooling processing is applied to each time step.

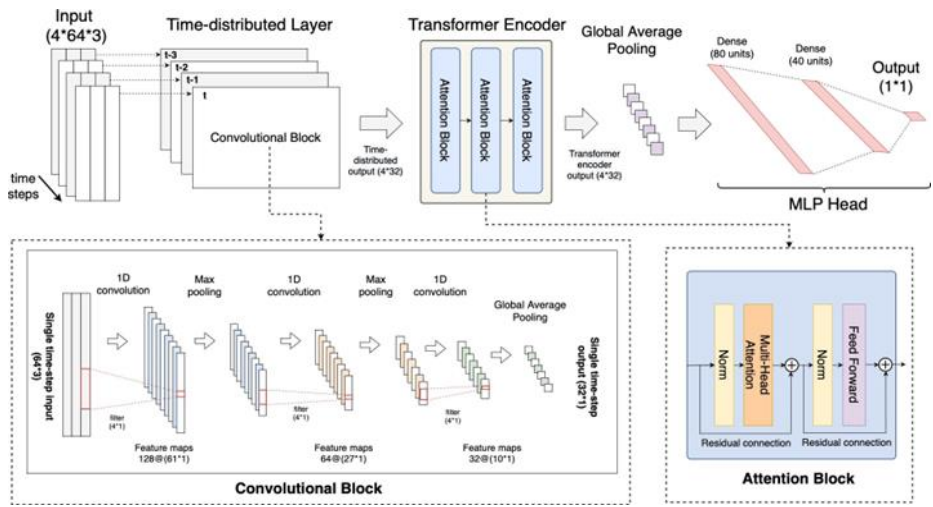

Figure 5:     Deep neural network with convolutional and Transformer blocks (FOG- Transformer).

The convolutional block is composed of an input layer (64 samples and 3 channels) and three 1-D convolutional layers with 128, 64, and 32 filters respectively, all of them with Rectified Linear Unit (ReLU) activation function and a kernel size of 4. The first two convolutional layers are connected to max-pooling layers with a pool size of 2, while the last convolutional layer is connected to a GAP layer. The output of the convolutional block has a size of 4*32, which corresponds to a sequence of 4 time steps with 32 features each.

2. Transformer encoder. It is responsible for analyzing the temporal information of the sequential data provided from the time-distributed layer. The encoder generates an attention-based representation with the capability to locate specific and relevant information across the whole (spectral) sequence. The encoder is composed of 3 Attention blocks, each with Multi-Head self-attention layers (Vaswani et al., 2017) (3 heads with a size of 32), normalization layers, and a Feed-Forward section with 1-D convolutional layers (16 filters, kernel size of 1, and dropout 0.25). The encoder employs residual connections and dropout as proposed in keras.io (2021). The output of the Transformer encoder is connected to a GAP layer to reduce the dimensionality of each time step to enable the connection to the classification (fully-connected) layers.

3. MLP Head: The last part of the FOG-Transformer is used for classification tasks. The MLP Head is composed of three fully-connected layers with 80, 40, and 1 neuron respectively. The first two layers use ReLU activation and dropout of 0.4, while the output layer employs a sigmoid activation to provide the probability in the classification of FOG (or Non-FOG) events.

### 3.5.2. FOG-Transformer Training and Hyperparameter Optimization

The training of the DNNs is an iterative process that repeats until finding an acceptable solution for a problem. In this process, the weights and biases of the network layers are updated using the back-propagation algorithm to minimize a loss function. The training of a DNN is difficult because of the large number of parameters to be adjusted in each layer (Glorot and Bengio, 2010).

Hyperparameters like the learning rate, batch size, the number of epochs or number of layers, and the layer's parameters remain constant during training. Thus, the accurate selection (optimization) of these hyperparameters has a great influence on the model performance and controls the training process in terms of computational processing.

In this work, the hyperparameter optimization of the FOG-Transformer model was performed with the hyperband method. The method presents a good trade-off between speed and performance in problems with high dimensionality space (Li et al., 2017). The FOG-Transformer was trained with the Adaptive moment estimation optimizer (ADAM) (Kingma and Ba, 2014) with a learning rate of $6*10^{-4}$, binary cross-entropy as loss function, a batch size of 512, and 150 as the maximum number of epochs. In addition, an early-stopping strategy was employed to prevent overfitting and unnecessary computing during training. This strategy stops the training when the loss value did not decrease for 7 continuous epochs. To apply this strategy the training data was subdivided into 80% for training (train–train) and 20% for validation (train–validation).

### 3.6. Evaluation Methodology for FOG Detection Algorithms
### 3.6.1. Leave-one-subject-out Cross-validation (LOSO CV)

In this study, the FOG detection algorithms were evaluated through LOSO CV. This evaluation method has been employed to develop generalized FOG detection systems that can be used with data from new patients that are not included during training. LOSO CV provides a subject-independent estimate of the performance for new subjects (Gholamiangonabadi et al., 2020), in contrast to the cross-validation methods (i.e., random k-fold) or the hold-out evaluation, which can employ data from the same subjects for training and evaluation, or data from specific subsets or selected subjects.

In LOSO CV, the data of all patients except one is used for training, while the data of the remaining patient is used for evaluating the model; then, this process is repeated for each patient. The performance of a model using LOSO CV can be analyzed for each subject, and the overall results can be calculated by averaging the partial results from each patient.

In addition, LOSO CV is more appropriate for the evaluation of data pro- cessed with sliding windows with overlap, to prevent signal segments to be shared between training and validation subsets (i.e., when using a random k- fold validation). In this study, the LOSO CV was performed using the data from 21 patients and the process was repeated six times for each subject. The reported results were computed by averaging the results of all evaluations.

### 3.6.2. Performance Metrics

To evaluate the performance of the proposed methods and those reproduced from the state-of-the-art approaches, several metrics were calculated and re- ported. For a binary classification problem (FOG or non-FOG), the results were expressed in terms of sensitivity, specificity, AUC of the receiver operating characteristic curve (ROC) (Hanley and McNeil, 1982), and EER (Jothilakshmi and Gudivada, 2016).

The sensitivity is the ratio of positives that are correctly identified, while specificity is the ratio of negatives that are correctly identified. These metrics were calculated using the number of true positives (TP), true negatives (TN), false positives (FP), and false negatives (FN). While the TP and the TN are the numbers of correctly identified positive and negative samples, the FP repre- sents the number of negative samples wrongly classified as positive, and the FN represents the number of positive samples wrongly classified as negative. The sensitivity and specificity were calculated using Eq. (1) and Eq. (2) respectively.

$$sensitivity = \frac{TP}{TP + FN} \qquad (1)$$

$$specificity = \frac{TN}{TN + FP} \qquad (2)$$

Because FOG detection algorithms can provide a probability value in their output, rather than a specific class, it is necessary to apply a classification threshold. For the comparison of the proposed approaches, the values of sen- sitivity and specificity have been obtained using the equal error rate threshold (EER threshold) (Freeman and Moisen, 2008).

In addition, the accuracy (number of correct predictions out of all data) and F-score (harmonic mean of the model's precision and sensitivity) were computed to evaluate the performance of the system when performing post-processing tasks for an overall analysis of FOG events. The accuracy and F-score were obtained using Eq. (3) and Eq. (4) respectively. The precision was computed using the Eq. (5).

$$accuracy = \frac{TP + TN}{TP + TN + FP + FN} \quad (3)$$

$$F - score = \frac{2 \cdot sensitivity \cdot precision}{sensitivity + precision} \quad (4)$$

$$precision = \frac{TP}{TP + FP} \quad (5)$$

### 3.7. FOG Detection Post-processing

For further analysis, a post-processing analysis was performed using the best approach for FOG detection. Since the output of the FOG detection algorithms is a probability (instead of a specific class FOG/Non-FOG), the effect of tuning the classification thresholds was evaluated using the F-score and the geometric mean (GM) between sensitivity and specificity (Section 3.7.1).

In addition, the performance in the detection of FOG episodes was analyzed (Section 3.7.2), besides the window-level FOG detection. Finally, a methodology for the analysis of the clusters of FOG episodes was proposed and evaluated (Section 4.6.3).

#### 3.7.1. Threshold Tuning

The tuning procedure on the classification threshold was performed taking as input the label and the continuous prediction score provided by the proposed model. The classification threshold was tuned in the range 0.2-0.8, and the GM and the F-score were evaluated while increasing the threshold value. Perfor- mance was compared for two threshold selection strategies, namely EER mini- mization and F-score maximization. The rest of the subsequent post-processing on FOG detection was performed for each of the threshold selection strategies.

#### 3.7.2. FOG Episode Detection

The analysis of FOG episodes and false FOG episode (FFE) was performed by analyzing groups of consecutive windows with the presence of freezing episodes. The analysis was based on the results of the window-based FOG detection pro- vided by the FOG-Transformer model. FOG Episode Detection and FFE de- tection were performed using the methodology described as follows.

FOG episode detection: When considering FOG episodes (i.e., group of consecutive windows labeled as FOG), the percentage of episodes detected and the proportion of FOG detected in each episode were computed. As for the former, an episode was considered detected if at least one window of labeled FOG inside that episode was correctly identified by the model. And for the proportion of FOG detected in each episode, the number of windows detected as FOG in each episode was divided by the total labeled FOG windows included in that episode. The analysis was performed both independently of the duration of FOG episodes and dividing episodes based on their duration. FOG episodes were divided into three groups, including episodes lasting less than 5s, episodes with duration in the range of 5-10s, and finally, episodes of duration longer than 10s.

False FOG episode detection: FFE were identified as groups of consecutive windows predicted by the model as FOG, while none of them were labeled as FOG. The percentage of FFE was computed as the number of FFE divided by the total number of predicted episodes. For each FFE, the distance from the nearest FOG was computed. This was done by performing four consecutive steps: Compute the number of windows between the beginning of the FFE and the end of the previous labeled FOG episode; compute the number of windows separating the end of the FFE and the beginning of the following labeled FOG episode; compute the minimum value between them; multiply the latter value by the sliding window size, which was set to 0.8s, to convert the number of windows into a time interval. Finally, the percentage of FFE far less than 5s and 10s from the nearest FOG was computed.

### 3.7.3. Clustering of FOG episodes

To analyze the performance of the proposed algorithm in detecting clusters of FOG episodes, the following processing was employed. First, the upper root- mean-square envelope of both the label vector and the binary prediction score was computed using a sliding window of 110 windows, corresponding to 1.5 min of data. Then, the Pearson correlation coefficient and the corresponding p-value were computed considering the envelopes of the labeled and predicted vectors. Finally, real FOG clusters were identified as the portion of data in which the labeled envelop exceeded the zero-value.

As for the predicted clusters, a threshold of 0.1 was selected to define FOG clusters. Correctly identified FOG clusters were defined as labeled clusters in which at least one window was predicted by the model as FOG. Predicted clusters in which none of the windows were labeled as FOG were defined as false clusters. The percentage of true/false clusters was computed by dividing the number of true/false by the total number of labeled/predicted clusters. Finally, duration and FOG content (i.e., percentage of windows predicted as FOG) for each cluster were computed, and the results were compared between true and false predicted FOG clusters. The Mann-Whitney U test was performed to verify whether those metrics were significantly different in the two populations.

## 4. Results and Discussion

In this section, the experiments and the results obtained from the different methods for FOG detection at window-level and post-processing analysis are reported and discussed. All the detection methods were evaluated through LOSO CV evaluation, the complete process was repeated six times to verify the variability in the results due to the stochastic processes in the training procedure.

### 4.1. Baseline Results

To make a baseline, the Mazilu features in conjunction with a Random Forest classifier with 100 estimators were evaluated. Table 2 presents the results in terms of sensitivity, specificity, AUC, and EER using LOSO CV. The results are presented for each subject included in the experiment.

Table 2: Results from the baseline model (Mazilu et al., 2012).

| Patient Index | Sensitivity | Specificity | AUC | EER(%) |
| --- | --- | --- | --- | --- |
| 1 | 0.844 | 0.833 | 0.928 | 16.3 |
| 2 | 0.708 | 0.695 | 0.771 | 30.5 |
| 3 | 0.899 | 0.904 | 0.960 | 9.6 |
| 4 | 0.786 | 0.803 | 0.898 | 19.7 |
| 5 | 0.824 | 0.821 | 0.905 | 17.9 |
| 6 | 0.839 | 0.832 | 0.931 | 16.8 |
| 7 | 0.855 | 0.855 | 0.933 | 14.5 |
| 8 | 0.851 | 0.857 | 0.914 | 14.3 |
| 9 | 0.776 | 0.782 | 0.879 | 21.8 |
| 10 | 0.865 | 0.886 | 0.943 | 13.4 |
| 11 | 0.831 | 0.819 | 0.904 | 18.1 |
| 12 | 0.776 | 0.762 | 0.846 | 23.8 |
| 13 | 0.844 | 0.837 | 0.930 | 16.3 |
| 14 | 0.912 | 0.915 | 0.971 | 8.5 |
| 15 | 0.898 | 0.907 | 0.961 | 9.3 |
| 16 | 0.851 | 0.857 | 0.929 | 14.3 |
| 17 | 0.876 | 0.877 | 0.940 | 12.4 |
| 18 | 0.885 | 0.887 | 0.954 | 11.3 |
| 19 | 0.852 | 0.839 | 0.942 | 16.1 |
| 20 | 0.759 | 0.774 | 0.870 | 22.6 |
| 21 | 0.861 | 0.861 | 0.933 | 13.9 |
| Average | 0.838 | 0.837 | 0.916 | 16.3 |

According to Table 2, the baseline method presents sensitivities and specificities ranging from 0.7 to 0.91. These results are lower than those reported in Mazilu et al. (2012) (0.995 in sensitivity and 0.999 in specificity) using the Daphnet dataset. These differences are expected because Daphnet considers a different number of sensors and the data collection includes activities performed under controlled conditions in contrast to the normal ADL used in this study.

The baseline reproduction shows an average AUC of 0.916 and an EER of 16.3. This method presents a good trade-off between ease of implementation and results, even considering that is expected a decrease in the performance metrics using LOSO CV methodology in comparison to k-fold cross-validation methods.

### 4.2. Evaluation of Different FOG Detection Methods

In this experiment, different classification algorithms were tested using their corresponding data representations. Table 3 shows a summary of the methods evaluated and the results in terms of sensitivity, specificity, AUC, and EER (%) obtained through LOSO CV.

Table 3: Performance of different methods for FOG detection.

| Method | Classifier | Sensitivity | Specificity | AUC | EER(%) |
| --- | --- | --- | --- | --- | --- |
| Mazilu et al. (2012) | Random Forest | 0.838 | 0.837 | 0.916 | 16.3 |
| Bikias et al. (2021) | CNN-MLP | 0.856 | 0.857 | 0.921 | 14.3 |
| Camps et al. (2018) | CNN-MLP | 0.863 | 0.863 | 0.938 | 13.7 |
| Sigcha et al. (2020) | CNN-LSTM | 0.871 | 0.871 | 0.939 | 12.9 |
| Present study | FOG-Transformer | 0.891 | 0.891 | 0.957 | 10.9 |

The results of the methods (Bikias et al., 2021; Camps et al., 2018; Sigcha et al., 2020) reproduced from the literature were achieved using the configuration (i.e., learning rate, number of epochs, batch size) that reported the best performance in the training and evaluation process. While the hyperparameters used in the FOG-Transformer are described in Section 3.5.2.

According to Table 3, the FOG-Transformer trained with the FFT plus 3 previous windows presents the best performance compared to other methods, by reaching a sensitivity, specificity of 0.891 each, an AUC of 0.957, and a reduction of the 5.4% in the EER in comparison with the baseline. The AUC of the FOG-Transformer increased from 0.939 to 0.957 in comparison with the best (CNN-LSTM) method reproduced from the related literature. With this dataset, a difference of 0.018 (1.8%) in AUC can be considered significant with p¡0.0005, according to Hanley's method (Hanley and McNeil, 1982).

The baseline method that uses a shallow ML algorithm presents a lower performance than the DL approaches with an AUC of 0.916 and an EER of 16.3%.

On the other hand, the results obtained with the DL methods based on CNN show similar performances in terms of sensitivity and specificity, while the addition of recurrent layers to CNN presents a slight increase in the performance and a corresponding reduction in the EER. According to these results, the method proposed in Bikias et al. (2021) presents a feasible approach for FOG detection with a minimal signal pre-processing, whereas the use of the spectral representation (FFT) with previous windows (Camps et al., 2018; Sigcha et al., 2020) can bring an additional improvement in the predictive performance of the DL based methods.

As shown in Table 3, the FOG-Transformer architecture presents a significant improvement over the best method reproduced for the related literature using the same number of previous windows and overlap settings (75%) proposed in Sigcha et al. (2020). These results suggest that the use of Transformers and attention blocks can improve the performance in FOG detection based on the analysis of adjacent windows and enables the development of DNNs without the use of recurrent layers to model sequential data. Also, the use of CNN blocks within a time-distributed layer seems to be a suitable mechanism for automatic feature extraction to feed recurrent layers (i.e., GRU or LSTM) or transformer blocks.

The addition of different positional encoding strategies before the Trans- former blocks did not present significant differences in the performance of the proposed FOG-Transformer architecture. The use of a trainable positional encoding shows slightly lower performance (AUC 0.950), as well as the use of a fixed positional encoding scheme proposed in Vaswani et al. (2017) (AUC 0.949).

Similar behaviors were reported using CNN layers before Transformers for im- age classification, showing slight differences in the accuracy among approaches and turning optional the use of positional embedding (Hassani et al., 2021).

### 4.3. Evaluation of the FOG-Transformer with Different Number of Previous Windows and Overlap

In a previous study (Sigcha et al., 2020), the addition of a larger number of previous windows using a CNN-LSTM model has shown a progressive increase in the overall performance in FOG detection, however, this approach presents a trade-off between performance and computational burden. Thus, the accurate selection of the windows size and overlap setting turns critical to reducing the computational burden produced for the feature extraction process, in particular when using sliding windows with a high overlap setting.

In this experiment, the performance of the FOG-Transformer with a different number of previous windows and overlap settings was evaluated. Table 4 presents the results of the performance of the FOG-Transformer model in terms of sensitivity, specificity, AUC, and EER using LOSO CV.

Table 4: FOG-Transformer performance at different overlap settings and previous windows.

| Overlap | No. of Previous Windows | Sensitivity | Specificity | AUC | EER(%) |
|---|---|---|---|---|---|
| 50% | 1 | 0.880 | 0.880 | 0.947 | 11.97 |
|  | 2 | 0.889 | 0.889 | 0.951 | 11.15 |
|  | 3 | 0.888 | 0.888 | 0.949 | 11.17 |
| 75% | 1 | 0.878 | 0.878 | 0.946 | 12.16 |
|  | 2 | 0.886 | 0.886 | 0.950 | 11.41 |
|  | 3 | 0.891 | 0.891 | 0.957 | 10.86 |

The results in Table 4 indicate a higher performance in terms of AUC by using the current and three previous windows with an overlap of 75%. However, the use of the current and two or three windows with an overlap of 50% show similar performances (AUC 0.951 and 0.949 respectively) than that obtained using two previous windows with an overlap of 75% (AUC 0.950). The use of a reduced overlap setting (i.e., 50% or less) can limit the number of windows to be analyzed in a given time interval, thus, reducing the computational burden produced by the feature extraction procedure without a significant decrease in performance. This situation presents opportunities for the development of accurate detection systems that can be used for long-term monitoring in real-life settings like ambulatory and home environments.

### 4.4. Comparison of the Performance and Number Trainable Parameters of the DL Models

To compare the performance and size of the DNNs, Table 5 describes the number of features used to feed the DL models, the number of trainable parameters, and its corresponding performance in terms of AUC.

Table 5: Performance, number trainable parameters and input shape of the DL models.

| Model | No. of features (input shape) | No.of Trainable Parameters | AUC |
|---|---|---|---|
| CNN-MLP(Bikias et al., 2021) | 384 (128*3) | 43,181 | 0.921 |
| CNN-MLP(Camps et al., 2018) | 384 (64*6) | 32,961 | 0.938 |
| CNN-LSTM (Sigcha et al., 2020) | 768 (4*64*3) | 288,993 | 0.939 |
| FOG-Transformer | 768 (4*64*3) | 87,825 | 0.957 |

As shown in Table 5, the reproduction of the deep architectures proposed in Camps et al. (2018) and Bikias et al. (2021) present a lower number of trainable parameters and a reduced number of input features in comparison with the CNN-LSTM and FOG-Transformer. However, the best results in terms of AUC were achieved with the CNN-LSTM model and the proposed FOG-Transformer using as input a 3-D data representation with 768 features. Although the FOG- Transformer presents a higher number of trainable parameters over the models proposed in Camps et al. (2018) and Bikias et al. (2021), the model exhibits a reduction of 201,168 parameters in comparison to the CNN-LSTM.

The results in Table 5 show a trade-off between performance and complexity among the DL approaches. However, the reduction in the number of trainable parameters in the FOG-Transformer could facilitate its implementation in real- time detection systems and stand-alone devices while enhancing the performance in FOG detection. According to these results, the subsequent experiments were performed with the FOG-Transformer model.

### 4.5. Results of the FOG-Transformer with Previous Windows Per Patient

For comparison purposes, Table 6 shows the results per patient obtained with the FOG-Transformer and the data representation based on the current and three previous windows (overlap 75%).

Table 6: Results of the FOG-Transformer with three previous windows per patient.

| Patient Index | Sensitivity | Specificity | AUC | EER(%) |
|---|---|---|---|---|
| 1 | 0.926 | 0.927 | 0.982 | 7.3 |
| 2 | 0.787 | 0.786 | 0.874 | 21.4 |
| 3 | 0.925 | 0.925 | 0.969 | 7.5 |
| 4 | 0.818 | 0.819 | 0.920 | 18.1 |
| 5 | 0.881 | 0.881 | 0.957 | 11.9 |
| 6 | 0.888 | 0.888 | 0.954 | 11.2 |
| 7 | 0.909 | 0.905 | 0.971 | 9.5 |
| 8 | 0.897 | 0.898 | 0.943 | 10.2 |
| 9 | 0.863 | 0.863 | 0.946 | 13.7 |
| 10 | 0.908 | 0.909 | 0.970 | 9.9 |
| 11 | 0.887 | 0.887 | 0.953 | 11.3 |
| 12 | 0.866 | 0.867 | 0.941 | 13.3 |
| 13 | 0.895 | 0.895 | 0.965 | 10.5 |
| 14 | 0.936 | 0.936 | 0.982 | 6.4 |
| 15 | 0.939 | 0.939 | 0.983 | 6.1 |
| 16 | 0.910 | 0.911 | 0.965 | 8.9 |
| 17 | 0.920 | 0.920 | 0.972 | 8.0 |
| 18 | 0.931 | 0.931 | 0.981 | 6.9 |
| 19 | 0.888 | 0.888 | 0.964 | 11.2 |
| 20 | 0.836 | 0.836 | 0.940 | 16.4 |
| 21 | 0.908 | 0.909 | 0.960 | 9.1 |
| Average | 0.891 | 0.891 | 0.957 | 10.9 |

According to Table 6, the AUC value is in the range of 0.874 to 0.982, with a mean AUC of 0.957, while the mean EER is 10.9%. The results show a high performance in FOG detection in the majority of the patients, however, the average result is affected by patients 2, 4, and 20. The variation in the results is expected because of the different freezing patterns and the low amount of freezing events found in these subjects.

When comparing the results per patient of the baseline (see Table 2) and the FOG-Transformer (see Table 6), increases of up to 10.3% in AUC (mean 4.1%; standard deviation 2.5%) and reductions of up to 10.5% in EER (mean 5.4%, standard deviation 2.3%) are observed. Comparison exhibits increases in AUC and reductions in EER for all subjects when using the FOG-Transformer; these results show a high predictive capability and the potential to predict FOG using data from new subjects. However, values as low as 1% increase in AUC have been observed in patients 3 and 14, which still present a challenge in the algorithmic area.

Despite partial results for specific subjects, the overall results suggest that the FOG-Transformer model presents a high generalization capability (model's capability to adapt properly to new, previously unseen data), avoiding the need to collect new data to train personalized models.

### 4.6. Results of the FOG Detection Post-processing

The results of the post-processing are presented in the next subsections. The results include the classification threshold tuning (Section 4.6.1), FOG event and FFE detection (Section 4.6.2), and the clustering of FOG episodes (Section 4.6.3). These results were obtained using the window-based predictions derived from the FOG-Transformer fed with the 3-D data representation with three previous windows.

### 4.6.1. Results of the Threshold Tuning

Fig. 6 reports the classification performance, in terms of GM between sensi- tivity and specificity, and F-score, for different values of the classification thresh- old (thr).

As shown in Fig. 6, as thr increases up to 0.8, the GM curve decreases from

0.91 to 0.76. The best performances in this metric were achieved using a low threshold setting. This trend is expected due to the unbalanced data that is intrinsic to this symptom.

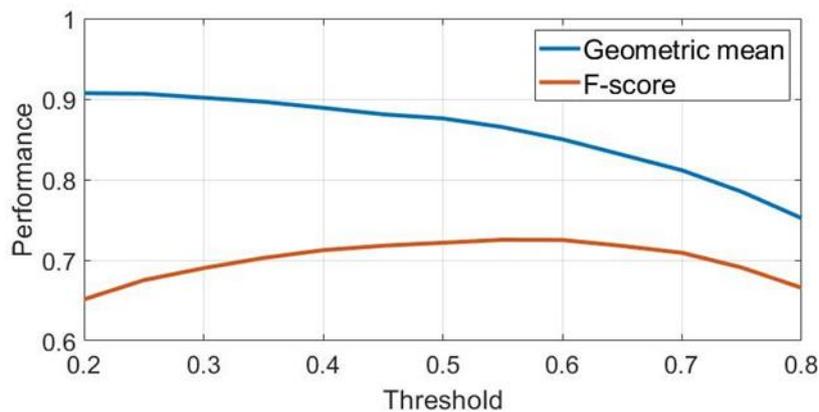

Figure 6: Classification performance using different threshold values.

As far as concerns the F-score, it increases until thr=0.6, where a maximum of 0.72 is observed, and then decreases for higher threshold values. However, while the F-score increases by only 1.2% in the interval 0.4-0.6, a reduction of GM by 3.9% is registered in the same interval. This suggest that thr =

0.4 represents a good compromise between GM and F-score. Compared to the EER-based approach, the F-score approach (thr = 0.4) improves F-score by 6.1%, with a reduction in GM by only 1.8%.

For further analysis, Table 7 reports the classification performance obtained using two different thr selection methods, namely EER minimization and the thr selected using the F-score method.

Table 7: Classification performance using different classification threshold criteria.

| Threshold | | Performance | | | | | |
|---|---|---|---|---|---|---|---|
| Criterion | Value | Accuracy | Sensitivity | Specificity | Geometric mean | Precision | F-score |
| EER minimization | 0.22 | 0.907 | 0.907 | 0.907 | 0.907 | 0.533 | 0.671 |
| F-score method | 0.40 | 0.929 | 0.842 | 0.939 | 0.889 | 0.617 | 0.712 |

Using the F-score method, accuracy (+2.2%), specificity (+3.2%), precision (+8.4%), and F-score (+4.1%) increase over the EER minimization. On the other hand, the EER minimization presents an increase in the geometric mean (+6.5%) and sensitivity (+1.8%). According to these results, both approaches show their applicability depending on the need for high precision and accuracy, or a balanced performance between sensitivity and specificity.

### 4.6.2. Results of the FOG Episodes Detection

When analyzing FOG episodes, the results for the EER threshold and the F- score method (in parentheses) are reported in the following. The 95.5% (91.2%) of episodes were detected by the algorithm, with an average proportion of 84.2% (74.6%) of FOG detected in each episode. The algorithm exhibited a different detection rate based on FOG episodes duration, as reported in Table 8. As FOG episodes duration increases, both detection rate and proportion of FOG detected in each episode increase. Moreover, the EER-based threshold method provided better performance than those obtained using the F-score approach.

Table 8: Detection rate of FOG episodes based on duration.

| FOG episodes duration (s) | <5s | | 5-10 s | | >10s | |
|---|---|---|---|---|---|---|
| Threshold criterion | EER | F-score | EER | F-score | EER | F-score |
| % FOG episodes detected | 92.4 | 86.0 | 99.3 | 98.6 | 100 | 100 |
| % FOG detected in each episode | 77.9 | 68.7 | 89.1 | 82.4 | 93.2 | 87.9 |

On the other hand, when analyzing FFE, 55.6% (44.8%) of the total episodes detected were found to be false positives, with a mean duration of 4s (3.9s), for the EER (F-score)–based threshold selection. From the total number of false episodes detected, 35.7% (40.1%) were found to be less than 5s distant from the nearest real FOG episode. Table 9 reports the proportion of false episodes detected, their duration, and the distance from the nearest FOG. Removing short detected FOG episodes present a slight improvement in F-score. More in detail, deleting episodes including 1, 2, 3 windows led to an improvement in F-score by 0.5% (0.4%), 1.2% (1.1%), 2.4% (1.6%), while reducing sensitivity by 0.2% (0.3%), 0.4% (0.6%), 0.6% (1.3%), for the EER (F-score) threshold selection strategy.

Table 9: False FOG episodes detection performance.

| Threshold | % FFE | FFE duration (s) | % FFE <5s far from FOG | % FFE <10s far from FOG |
|---|---|---|---|---|
| EER | 55.6 | 4 ± 0.9 | 36.3 | 47.8 |
| F-score | 44.8 | 3.9 ± 0.8 | 40.1 | 52.0 |

As shown in Table 9, using the F-score threshold selection strategy led to the reduction of over 10% in FFE, with a larger percentage of FFE located close to real FOG episodes. This latter result is important for two reasons. In the case of a real-time implementation of the algorithm, false episodes detected before the actual onset of FOG may be beneficial for triggering some sort of cueing system. Conversely, when using the classification algorithm for offline processing of daily data, false episodes detected near the real FOG do not significantly affect the performance of the algorithm, in terms of accumulations of FOG episodes. This latter point is discussed in detail in the next section.

### 4.6.3. Results of the Clustering of FOG Episodes

When analyzing clusters of FOG episodes, the envelope of detected FOG episodes is strongly correlated with that of labeled FOG episodes, as shown in Fig. 7. Table 10 reports the Pearson correlation coefficient (r), the percentage of true clusters detected, the percentage of FOG detected in each cluster, and finally, the percentage of false clusters detected.

Table 9: False FOG episodes detection performance.

| Threshold | r (p-value) | % clusters detected | % FOG detected in each cluster | % false clusters detected |
|---|---|---|---|---|
| EER | 0.89 (<0.001) | 98.6 | 99.8 | 36.3 |
| F-score | 0.91 (<0.001) | 96.9 | 99.9 | 21.4 |

As can be observed from the table, the percentage of clusters detected is slightly larger using the EER threshold (+1%), while the percentage of false clusters detected is significantly lower for the F-score threshold (-11.4%). Pear- son correlation coefficient and the percentage of FOG detected in each cluster are similar for the two approaches.

The Mann-Whitney U-test demonstrated that true and false predicted clus- ters were different for duration and FOG content (p¡0.001), with false clusters being shorter (average: 1.42 min vs 6.17 min and 1.45 min vs 5.65 min for the EER and F-score approach, respectively) and including smaller FOG amount (average: 8.9% vs 27.6% and 8.6% vs 31.5% for the EER and F-score approach, respectively) than true clusters, as can be observed in Fig. 7. Clustering of FOG episodes represents an effective tool for summarizing the FOG distribu- tion throughout the day of PD patients. In fact, the accumulation of FOG episodes during specific periods of the day are strongly related to the OFF ther- apy condition, when the pharmacological treatment is no longer effective (Borzì et al., 2021; Suppa et al., 2017). Thus, results from clusters of FOG episodes may provide neurologists with useful information for scheduling proper therapy adjustments.

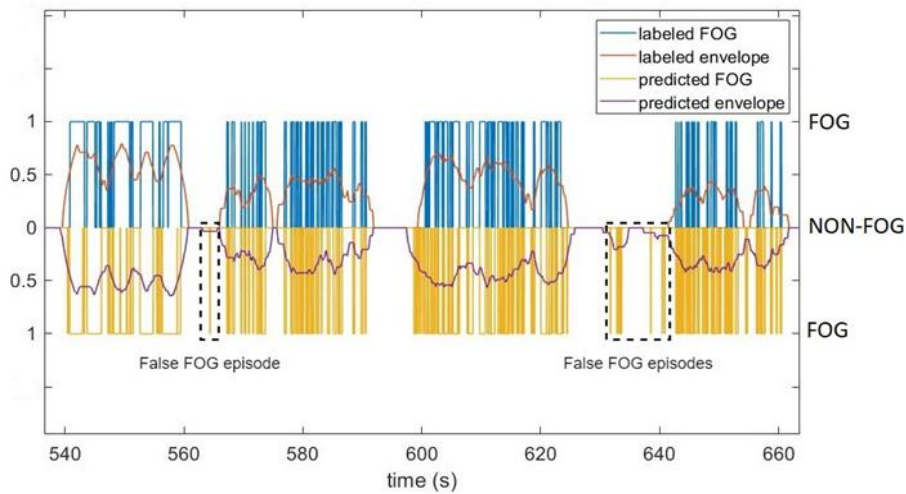

Figure 7: Labels and detected FOG episodes, together with computed envelops.

5.  **Conclusions**

In this paper, approaches for FOG detection using a single triaxial accelerometer (from a body-worn IMU) and FOG-Transformer networks have been evaluated. The data employed to evaluate the proposed methods were collected in the patients' homes where it is expected an increase in the occurrence of FOG events (Nieuwboer et al., 1998). The evaluation of the models was carried out with a LOSO CV to evaluate the performance of a FOG detection model generalizable for data from new subjects.

The best results were obtained with the FOG-Transformer model that included a convolutional block (for feature extraction from the triaxial accelerometer spectra) and a Transformer encoder with attention blocks to model time dependencies. A significant improvement in the performance of FOG detection was obtained using a 3-D data presentation that considers up to three previous spectral windows, by taking advantage of the temporal phases (movement pat- terns) that appear before the occurrence of a FOG event (Borzì et al., 2021; Cupertino et al., 2022). Moreover, the use of CNN blocks (within time-distributed layers) seems to be a suitable method for automatic feature extraction from spectral data to feed the Transformer blocks.

When comparing the baseline (hand-made features and a Random forest classifier) with the FOG-Transformer, an increase in AUC of 4.1% and a reduction of 5.4% in EER were achieved. The proposed FOG-Transformer architecture shows to be a suitable method for FOG detection without the need of using recurrent layers to model temporal dependencies in the data. Also, a significant reduction in the number of the trainable parameter is observed in comparison with the best method (CNN-LSTM (Sigcha et al., 2020)) reproduced from the related literature, thus reducing the computational complexity to support the implementation of this algorithm in long-term monitoring systems.

In the experiments made in this study, the FOG-Transformer has shown a capability of outperforming FOG detection methods based on convolutional layers (Bikias et al., 2021; Camps et al., 2018) and methods based on the combination of convolutional and recurrent layers (Sigcha et al., 2020). In addition, the results show the feasibility of using a single sensor for FOG detection. The use of a low number of sensors or devices should be considered to simplify the usability and increase the acceptability of systems intended for use in free-living or ambulatory settings (Rodríguez-Martín et al., 2017). The detailed post-processing performed on predicted FOG episodes revealed an excellent performance of the present algorithm in the detection of FOG episodes, while false episodes were located near real FOG. Finally, clustering of FOG episodes could be an effective tool for monitoring FOG in non-supervised environments during daily life. This can provide relevant information regarding the timing and the duration of FOG episodes accumulation, of fundamental importance for a proper schedule of pharmacological treatments and long-term monitoring.

Although the use of FOG-Transformer shows an enhancement in the FOG detection performance, the use of reduced window size is still a requirement for the development of algorithms with low latency to be used as part of a cue system to reduce the occurrence of FOG events without interrupting patients' daily life. According to the results, the use of a data presentation based on the current and two or three previous windows with an overlap of 50% seems to be a good trade-off between performance and computational burning, while maintaining a short latency derived from the window length and the computation time.

The main contributions of this paper are the implementation and evaluation of novel approaches based on Transformer and convolutional networks for FOG detection. Also, a methodology for FOG episode analysis and clustering is proposed with the aim of improving the outcomes of systems intended for long-term FOG analysis in ambulatory and free-living settings. The outcomes of these methods could be used in conjunction with standard gait parameters to provide reliable indicators that could be integrated into routine care, the implementation of gait assistance systems, and the remote assessment of PD patients.

Future work in algorithm optimization could focus on the contextualization of activities, thus limiting the burden of complex algorithms to operate only during the sections of interest, i.e., automatic activation of FOG detection algorithms only during walking. Also, a FOG prediction (Borzì et al., 2021;

Naghavi and Wade, 2021) should be addressed by improving the predictive power and the efficiency of detection methods to support the implementation of a robust cueing system.

In addition, the improvement in the precision of the detection models should be addressed to reduce the false-positive rate observed in the experiments. The implementation of novel pre-processing techniques or novel end-to-end prediction models could provide mechanisms for the implementation of accurate real-time cueing systems that help reduce the occurrence of FOG episodes and their consequent falls.

**Declaration of competing interest**

The authors declare that they have no known competing financial interests or personal relationships that could have appeared to influence the work reported in this paper.

**CRediT authorship contribution statement**

Luis Sigcha: Conceptualization, Methodology, Software, Investigation, Vi- sualization, Writing - Original Draft. Luigi Borzì: Conceptualization, For- mal analysis, Investigation, Software, Validation, Writing - Original Draft. Ig- nacio Pavón: Formal analysis, Project administration, Writing - review & editing. Nelson Costa: Funding acquisition, Supervision, Writing - review & editing. Susana Costa: Validation, Formal analysis, Writing - review & editing. Pedro Arezes: Resources, Supervision, Writing - review & editing. Juan Manuel López: Resources, Supervision, Writing - review & editing. Guillermo De Arcas: Funding acquisition, Project administration, Writing - review & editing.

**Acknowledgments**

This work was supported by "Tecnologías Capacitadoras para la Asisten- cia, Seguimiento y Rehabilitación de Pacientes con Enfermedad de Parkinson". Centro Internacional sobre el envejecimiento, CENIE (código 0348 CIE 6 E) Interreg V-A Espan˜a-Portugal (POCTEP); FCT- Funda¸c˜ao para a Ciˆencia e Tecnologia within the R&D Units Project Scope: UIDB/00319/2020; and the Grupo de Investigación en Instrumentación y Acu´stica Aplicada (I2A2). ETSI Industriales. Universidad Politécnica de Madrid.

The authors would like to thank Technical Research Centre for Dependency Care and Autonomous Living (CETpD) for sharing the data used in this study.